\documentclass[11pt]{article}
\usepackage{coling2020}
\usepackage{times}
\usepackage{url}
\usepackage{latexsym}
\usepackage{algorithm}
\usepackage{amsmath}
\usepackage{graphicx}
\usepackage[noend]{algpseudocode}

\usepackage{booktabs}

\usepackage{subcaption}
\usepackage{subfloat}
\colingfinalcopy 

\title{syrapropa at SemEval-2020 Task 11: BERT-based Models Design For Propagandistic Technique and Span Detection}

\author{Jinfen Li \\
  College of Arts and Sciences,\\Syracuse University\\
  {\tt jli284@syr.edu} \\\And
  Lu Xiao \\
  School of Information Studies,\\ Syracuse University \\
  {\tt lxiao04@syr.edu} \\}

\date{}

\begin{document}
\maketitle
\begin{abstract}
This paper describes the BERT-based models proposed for two subtasks in SemEval-2020 Task 11: \textit{Detection of Propaganda Techniques in News Articles}. We first build the model for Span Identification (SI) based on SpanBERT, and facilitate the detection by a deeper model and a sentence-level representation. We then develop a hybrid model for the Technique Classification (TC). The hybrid model is composed of three submodels including two BERT models with different training methods, and a feature-based Logistic Regression model. We endeavor to deal with imbalanced dataset by adjusting cost function.
We are in the seventh place in SI subtask (0.4711 of F1-measure), and in the third place in TC subtask (0.6783 of F1-measure)  on the development set.
\end{abstract}

\section{Introduction}
\blfootnote{
    
    \hspace{-0.65cm}  
    This work is licensed under a Creative Commons 
    Attribution 4.0 International License.
    License details:
    \url{http://creativecommons.org/licenses/by/4.0/}.
}
Propaganda exists in social media content and can subvert and distort public deliberation \cite{farkas2018disguised}. Natural language processing (NLP) researchers develop computational techniques that automatically detect propaganda in the content.\par
In this paper, we develop two systems for two subtasks in the SemEval-2020 Task 11: (1) Span Identification (SI) (2) Technique Classification (TC) in News Articles respectively. The SI subtask focus on identifying fragments in a given plain-text document which contain at least one propaganda technique; while TC subtask aims to classify the applied propaganda technique given a propagandistic text fragment \cite{DaSanMartinoSemeval20task11}.\par
For SI, we interpret the task as to detect a span in a context, which is the smallest detect unit. We base our system on SpanBERT \cite{joshi2020spanbert}, and combines three jointly trained classifiers: sentence, start, and end classifiers. Specifically, the start and end classifiers detect the start and end position of the span, while the sentence classifiers provide sentence-level information for the start and end classifiers. For TC, we come up with a hybrid system combining the two BERT models with different training methods, a Logistic Regression model, and extra rules to classify a propagandistic fragment into one of the 14 techniques. \par
For SI, we find that the segment approach of a context affects the result. While for TC, emotion features extracted from NRC lexicon are not effective to distinguish 14 classes. However, features such as text length, TF-IDF, occur times in a document, superlative form, question words, hashtags and supplement are useful in distinguishing different propaganda techniques. Overall, we are in the seventh place in SI subtask (0.4711 of F1-measure), and in the third place in TC subtask (0.6783 of F1-measure) on the development set.

\section{Related Work}
It is believed that news media plays an active and major role in producing and distributing propaganda and there can be a tool box that helps detect propaganda in news articles \cite{zollmann2019bringing}. \newcite{EMNLP19DaSanMartino} annotated 18 propaganda techniques in news articles. The researchers later merged the similar underrepresented techniques, resulting in 14 techniques \cite{DaSanMartinoSemeval20task11}.\par
The model trained on the sentence level task (whether the input sentence is propagandistic) is found to be effective for the fragment level task (detect the propagandistic fragment). \newcite{gupta2019neural} designed a multi-granularity and multi-tasking neural architectures to jointly perform both the sentence and fragment level propaganda detection. \newcite{da2019fine} designed a multi-granularity neural network that includes document-level, paragraph-level, sentence-level, word-level, subword-level and character-level task to detect fragments in news articles. Different from their studies, our SI model utilizes a sentence-level classifier which predicts whether a context contains a span or not, and embeds the hidden representations from it into the other two classifiers. We also explore different approach of concatenation of other higher representations.\par
Previous researches explore the effectiveness of hybrid model. For instance, \newcite{al2019justdeep} use several submodels including  BiLSTM, XGBoost, BERT model to predict if a sentence is propagandistic. Our TC model is also an hybrid model of three submodels (BERT, cost\_BERT, LR\dag\ddag) that combine the partial results based on their learning capacity of different categories. We compare the unweighted and weighted cost function in our approach and find out that the latter outperforms on the minority classes such as \textit{Whataboutism/Straw Men/Red Herring, Thought-terminating Cliches and Bandwagon/Reductio ad hitlerum}.\par

\section{SI System Description}
\subsection{Context Segmentation} \label{sec:context}
In the news article dataset provided by \newcite{DaSanMartinoSemeval20task11}, the start character index and end character index of each span is annotated in a news article. A span may be part of a sentence or include up to five sentences. The segmentation of the context is essential in the SI task as our goal is to detect a span in one context. In order to detect as many spans as possible in one news article, we first split the article into sentences. We then merge the overlapped spans into one span, for the reason that a longer span improves the recall when we only predict one span in a context.
We define two different contexts using the following strategies:

\begin{enumerate}
    
    \item \textbf{Mini Context} 
    For one sentence,\\
    \textbf{a)} if there are multiple merged spans: we extend the context from the span (including the merged one) to the left and right side until it meets the boundary of other spans. As shown in Figure \ref{fig:context}, for sentence 1, span 3 combining span 1 and 2, is within context 1; while span 4 is within context 2. \\
    \textbf{b)} if there is no span or one span in a sentence: then the sentence itself is one context. \par
    \begin{figure}[htbp]
	\centering
	\includegraphics[width=3in]{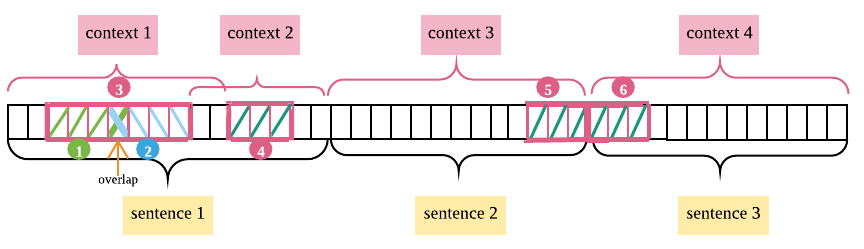}  
	\caption{Context segmentation. The red color represents the merged spans and the green and blue colors represents the unmerged spans.} 
	\label{fig:context}   
\end{figure}

\begin{figure*}[htbp]
	\centering
	\includegraphics[width=\textwidth]{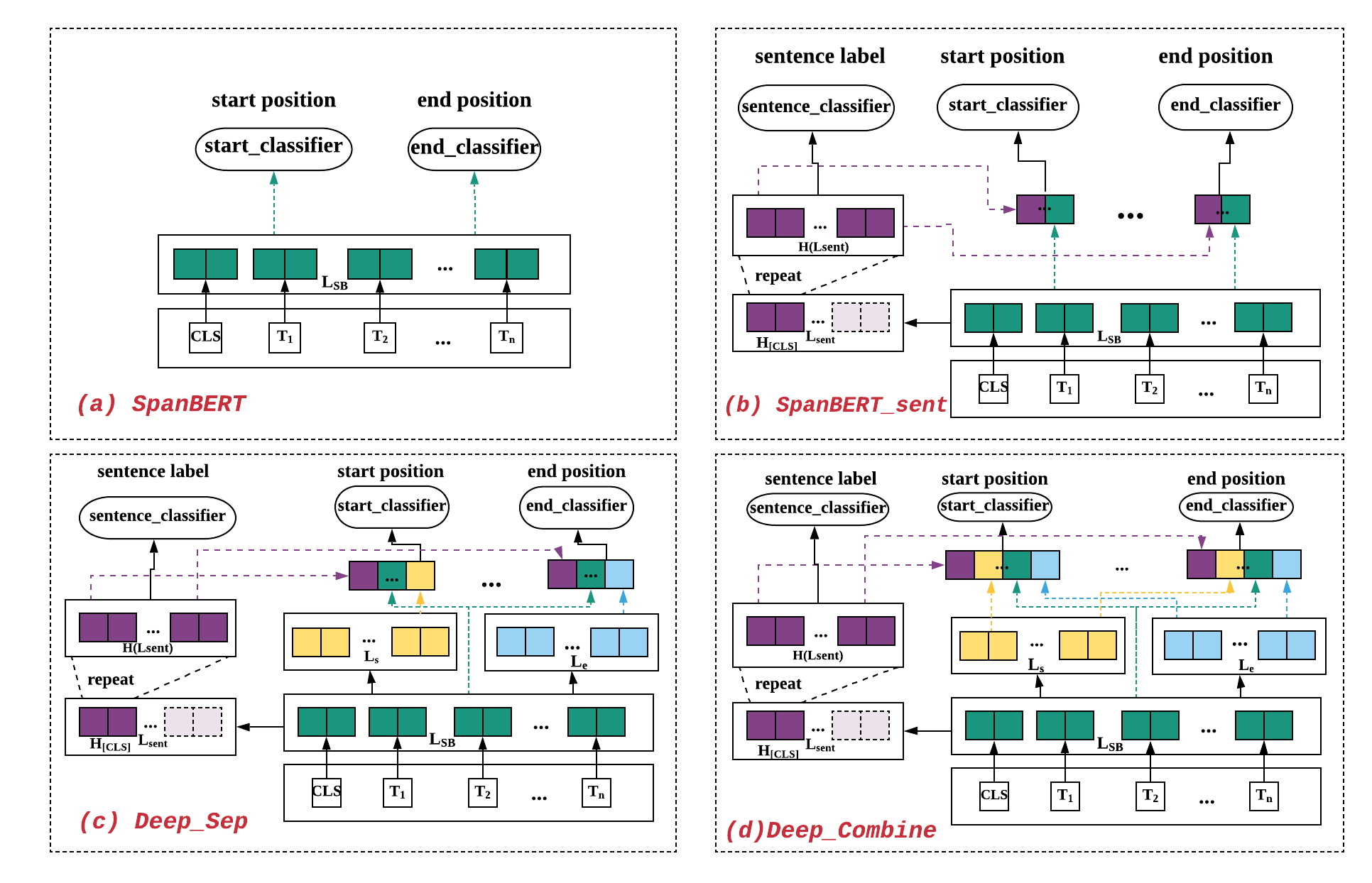}  
	\caption{The architecture of the SpanBert models (a),
and of our proposed modification networks (b-d).} 
	\label{fig:model}   
\end{figure*}
    In cases where a span covers across several sentences, it is split into multiple shorter spans at the sentence boundary. As shown in Figure \ref{fig:context}, the long span across sentence 2 and sentence 3 is splitted into span 5 and 6. The advantage to segment in this way is that it includes all the spans in training dataset and it has least noise in a context since there is only one span in a context, and the context is not too long for training. However, some contexts sacrifices their semantics integrity as they are just part of one sentence. We refer to it as ``mini context''.
    \item \textbf{Sentential Context} For the sentences containing multiple spans, we only keep the longest one and ignore the others. We refer to it as ``sentential context''.
\end{enumerate}

While for the development and test dataset, we detect the spans in each sentence.
\subsection{Our Model}
We based our model on the pre-trained \textit{SpanBERT\_base} model and modify the top layers. The overall architectures of SpanBERT is shown in Figure \ref{fig:model}-a, and our proposed models are shown in \ref{fig:model}-(b-d).

\subsubsection{Start and End Classifier}
We add two separate linear layers $L_s^o$, $L_e^o$ on top of SpanBERT and fine-tune it. The $L_s^o$ layer outputs the probability of each token being the start boundary of the span and the $L_e^o$ layer outputs that of each token being the end boundary of the span.
\subsubsection{Sentence Classifier}
The sentence classifier aims to classify whether a context contains a span, and the context is propagandistic if so. We come up with a layer $L_{sent}^k$ to capture sentence-level feature. As shown in Figure \ref{fig:model}-b, after feeding the hidden representations from the last layer of SpanBERT to $L_{sent}^k$, we only keep the feature of the first token ``[CLS]'', $H_{[CLS]}$ as it represents the whole context. We repeat $H_{[CLS]}$ to the number of tokens in the context and concatenate them to the hidden representation of each token. 
In addition, we feed $H(L_{sent}^k)$ into the output layer $L_{sent}^o$ for binary classification.
\subsubsection{Concatenation Layer}
 As shown in Figure \ref{fig:model}-c (\textbf{\textit{Deep\_Sep}}), the output layer of the above three classifiers directly accepts the SpanBERT hidden representation, we deepen our model by adding the layers $L_s^{k + 1}$,$L_e^{k + 1}$ to $L_s^{k}$,$L_e^{k}$, respectively, which extract a higher-level representation. In addition, we keep the residual connection from SpanBERT hidden layer to the deepened layers, i.e., $L_s^o$ and $L_e^o$. Therefore, $L_s^o$ and $L_e^o$ accepts concatenated representations from SpanBERT, $L_{sent}^k$ and $L^{k + 1}$. The two separate concatenated representations is shown in Equation \ref{eqa:4} and the overview of architecture is shown in \ref{fig:model}-c.
\begin{equation}\label{eqa:4}
    \begin{array}{l}
{H_s} = [{H_{SpanBERT}};H(L_s^{k + 1});H(L_{sent}^k)],\\
{H_e} = [{H_{SpanBERT}};H(L_e^{k + 1});H(L_{sent}^k)]
\end{array}
\end{equation}

As shown in Figure \ref{fig:model}-d (\textbf{\textit{Deep\_Combine}}), the above concatenation layer generates two separate representations to feed into start and end classifiers respectively. We find that in a news article, it is more likely that one start boundary maps only one end boundary and vice versa. In order for both start classifier and end classifiers to adopt information from each other, we concatenate the output of $L_s^{k + 1}$ and $L_e^{k + 1}$, together with the hidden representations from SpanBERT and $L_{sent}^k$, i.e., ${H_{combine}}$ in Equation \ref{eqa:3}.  ${H_{combine}}$ is then fed into start and end classifiers. The architecture is shown in Figure \ref{fig:model}-d.

\begin{equation}\label{eqa:3}
    {H_{combine}} = [{H_{SpanBERT}};H(L_s^{k + 1});H(L_e^{k + 1});H(L_{sent}^k)]
\end{equation}

\subsubsection{Our Loss Function}
We assign weights to different class (\textbf{\textit{cls\_weighted}}): we jointly train sentence, start and end classifiers in our model. The objective of the sentence classifier is the binary cross entropy loss.
\begin{equation}\label{eqa:1}
    L{_{sent}}(y,\hat y) =  - \sum\limits_i^N {{y_i}\log ({{\hat y}_i})} 
\end{equation}

As for our start and end classifiers, we adopt the multi-class cross entropy loss function. Because the proportion of the context with span (minority class) and without span (majority class) is imbalanced, we assign a weight to the minority class. The equation is give in Equation \ref{eqa:2}, where $w$ denotes the weight. Considering different convergence speed of the loss of three classifiers, we design the total loss function as Equation \ref{eqa:ave}. We combine the results of the best start index and the best end index, i.e., $span({I_s},{I_e}|{I_s})$ and $span({I_s}|{I_e},{I_e})$ in the prediction process, where $I$ is the boundary index.
\begin{equation}\label{eqa:2}
    L{_{start/end}}(y,\hat y) = \left\{ \begin{array}{l}
 - \sum\limits_i^N {{y_i}\log ({{\hat y}_i}),{y_{sent}} = 0} \\
 - \sum\limits_i^N {{y_i}\log (\textbf{w} \cdot {{\hat y}_i})} ,{y_{sent}} = 1
\end{array} \right.
\end{equation}



\begin{equation}\label{eqa:ave}
{L_{total}} = {\alpha _{sent}}{L_{sent}} + {\alpha _{start}}{L_{start}} + {\alpha _{end}}{L_{end}}
\end{equation}




\section{TC System Description}
Our system includes three individual sub-models combined together with extra rules, which outperforms any of the sub-models itself.

\subsection{Polymorphic BERT}
Bidirectional Encoder Representation Transformer (BERT) is a model based on the bidirectional transformer to embed more context information from left to right and from right to left. In order to incorporate features of emotion, we come up with the emotion representations and concatenate them with the default representation in BERT. To deal with the imbalanced dataset, we explore the performance of assigning different weights to different classes in the cost function.

\subsubsection{Embedding Emotion Feature} 
Emotional appeal is an important strategy used in propaganda techniques \cite{DaSanMartinoSemeval20task11}. \cite{li2019detection} found that features of emotions can be good indicators of a propagandistic and non-propagandistic fragment in news articles. We explore whether emotion features can help in the identification of the type of propaganda technique in the fragment.  \par
We use emotion lexicon NRC Affect Intensity Lexicon provided by \cite{mohammad2017wassa}, which contains the affect intensity of words in categories such as anger, disgust, joy, negative, positive, etc.  In order to utilize the pretrained uncased BERT-Base model, of which the hidden size is 768 \cite{devlin2018bert}, we come up with an emotion embedding table ${E_e}$ with the same hidden size, of which each row is randomly initialized except that the first ten values are the affect intensity score in the lexicon. In other words, we add the emotion representation to BERT (i.e., \textbf{emo\_BERT}) that typically includes three types of embedding: word embedding, position embedding, and segment embedding.

\subsubsection{Solving Class Imbalance}\label{sec:imb}
Because the training corpus is highly imbalanced (please check Appendix), we adjust the cost function (i.e., \textbf{cost\_BERT}). We first use the cross-entropy loss where $x$ is the softmax output and $y$ is the onehot encoding of the label and the ${t^{th}}$ element is the target class $y_t$ (Equation is shown in \ref{eqn:eq1}). We then multiple the cross-entropy loss with the weight of the target class $w_{{y_t}}$, where $w_{{y_t}}$ is the reciprocal of the frequency of $y_t$ in training dataset (Equation is shown in \ref{eqn:eq2}). With the modification of the cost function, the model will punish more on the mislabeled minority class (intuitively a more ``important'' class), such as \textit{Bandwagon,Reductio ad hitlerum, Thought-terminating,Cliches} and \textit{Whataboutism,Straw Men,Red Herring}.

\begin{equation}\label{eqn:eq1}
    loss(x,y) =  - \log (\frac{{\exp ({x_{{y_t}}})}}{{\sum\nolimits_j {\exp ({x_{{y_j}}})} }})
\end{equation}

\begin{equation}\label{eqn:eq2}
    \begin{array}{l}
weight\_loss(x,y) = {w_{{y_t}}} \cdot loss(x,y), \;\;\;\;
\text{where}\;\;\; {w_{{y_t}}} = \frac{{\sum\nolimits_j {{c_{{y_j}}}} }}{{{c_{{y_t}}}}}
\end{array}
\end{equation}

\subsection{Sub-Model:LR\dag\ddag}\label{sec:fea}
Our hybrid model combines the partial results generated by the sub-models including the typical BERT (\textbf{BERT}), the BERT trained with the cost-weighted function (\textbf{cost\_BERT}) and the Logistic Regression (\textbf{LR\dag\ddag}) introduced in this section.
We extract two continuous features:\textbf{Length, TF-IDF}; and several Boolean features including \textbf{Repetition, Superlative, Whatabout, Doubt, Slogan} and \textbf{Supplement}.
\begin{itemize}

    \item \textbf{Length} We found that the text length of most fragments in some categories tend to be shorter than those in the other categories. We use the text length as our baseline feature.
    \vspace{-1.5ex}
    \item \textbf{TF-IDF} We use TF-IDF values \cite{jones2004statistical} to enrich the dimension of features.
    \vspace{-1.5ex}
    \item \textbf{Repetition} The Repetition feature is a Boolean feature and is True if the fragment occurs more than four times in an article.
    \vspace{-1.5ex}
    \item \textbf{Superlative} The technique of \textit{ Exaggeration,Minimisation} utilize words in superlative format (e.g., ``largest'', ``best'', ``greatest'') to exaggerate or minimize some facts. The Superlative feature is a Boolean feature and is True if the fragment contains words in superlative form.
    \vspace{-1.5ex}
    \item \textbf{Whatabout} Just as the name of \textit{Whataboutism} tells, we detect whether the fragment starts with phrase ``what about''.
    \vspace{-1.5ex}
    \item \textbf{Doubt} Fragements that use \textit{Doubt} technique are likely to start with auxiliary words (e.g., has, is, do) or modals (e.g., can) or question words (e.g., why, what). With this Boolean feature we consider the signal in the classification.
    \vspace{-1.5ex}
    \item \textbf{Slogan} Fragments in \textit{Slogan} class contains words that start with hashtags (e.g., \#NeverAgain, \#StopTheSynod), or start with ``we will'' (e.g., ``we will serve the Lord''). The TRUE value of this Boolean feature means that the input fragment starts with a hashtag or ``we will''.
    \vspace{-1.5ex}
    \item \textbf{Supplement} \textit{Red Herring} technique introduces material, irrelevant to the focal issue, so as to divert people's attention away from the points made \cite{EMNLP19DaSanMartino}. Some fragments are encompassed by a pair of brackets like ``(who Kennedy admired)'', ``(Faber was nominated by President George H.W.Bush.)'', acting as a supplement. Some fragments use the ``who clause'' such as ``who is ...''. We view these linguistic expressions as a supplement to the sentence. This feature represents whether the the fragment is a supplement.
\end{itemize}

\subsection{Rule-based Correction and Reinforcement}\label{sec:rule}
After the prediction by the hybrid model, we apply simple syntactic rules to correct the mislabeled instances. Specifically, we compile rules based on part of speech tag (aka., POS tag) as follows.\par
For a fragment that is predictd to contain \textit{Repetition} but its occurrences is less than three times in the article, if its POS tag sequence contains (`NN',`NN') or (`NN',`NNS') or (`NNS'), it is corrected as \textit{Name Calling,Labeling}; if it is (`JJ') or (`NN'), it is corrected as \textit{Loaded Language}. Our experiment shows that this approach outperforms the alternative – to include this POS sequence as a feature in the Logistic Regression model. This is likely because fragments under the other categories may contain such POS tag sequences as well, adding noise to the classification. 

\section{Experimental Setup}
We use the training, development and test datasets provided by \cite{DaSanMartinoSemeval20task11}, which contains news articles from around 50 news outlets.
For SI, the evaluation function gives credit to partial matching between two spans \cite{DaSanMartinoSemeval20task11}. 
We base our model on the pretrained cased SpanBERT-Base model and fine-tuned them on development dataset with the following configuration: sequence length of $128$, learning rate of $1e-5$, batch size of $4$. We choose 64 for the hidden size of $L_s^{k + 1}$ and $L_e^{k + 1}$. In addition, ${\alpha _{sent}}, {\alpha _{start}}, {\alpha _{end}}$ equal to 0.25, 0.5 and 0.5.
For TC, we evaluate our model on by micro-averaged F1-measure. We base our BERT model on the pretrained uncased BERT-Base model and fine-tuned with the following configuration: sequence length of $128$, learning rate of $1e-5$, batch size of $4$. We utilize the solver of LBFGS, penalty of l2, C of $1.0$ and ``balanced'' mode in Logistic Regression.
\section{Results}\label{sec:results}
\subsection{Results of SI }
We outline the performance of a set of models in Table \ref{tab:result}.
For the models trained on sentential context, the adoption of an extra sentence classifier (\textit{SpanBERT\_sent}) outperforms the base SpanBERT (\textit{SpanBERT}). Our start and end classifiers, adopting the separate concatenated representation (\textit{Deep\_Sep}) and the combine concatenated representation (\textit{Deep\_Combine}), perform better comparing using the shallow representations. The decrease in F1-score of \textit{Deep\_Combine} implies that the start and end classifiers are conceptually equal and the boundary is not dependent to each other. 
To deal with the imbalanced dataset, we come up with the strategy to assign weights to different classes. \textit{cls\_weighted\dag} improves around 0.02 comparing to the unweighted model.\par
In contrast to these models trained on sentential context, the models trained on mini context mostly outperform them including \textit{SpanBERT, SpanBERT\_sent, Deep\_Sep} and \textit{Deep\_Joint}. \textit{cls\_weighted\dag} and \textit{loss\_weighted\dag\ddag} achieve similar F1-score with those trained on sentential context. This implies that remaining as many annotated minority classes as possible is essential when training on a small size of data. Our current model is not strict on semantics integrity. Also, while our strategy of identifying the sentential context ratains semantics integrity for the contexts, it loses some gold-labeled spans in the training dataset.
Lastly, we combine \textit{cls\_weighted\dag\ddag} trained both on sentential and mini context, which achieves 0.47108 of F1-score (SpanPro in Table \ref{tab:result}).

\begin{table*}[htbp]
  \centering
  \caption{Performance of each model. \dag represents the inclusion of Deep\_Sep into the model.\ddag~ represents the inclusion of cls\_weighted.}
    \begin{tabular}{lcccccc}\toprule
      & \multicolumn{3}{c}{sentential context} & \multicolumn{3}{c}{mini context} \\\midrule
    Model & F1 & Precision & Recall & F1 & Precision & Recall \\\midrule
    SpanBERT & 0.44576 & 0.39232 & 0.51604 & 0.44761 & 0.41362 & 0.48768 \\
    SpanBERT\_sent & 0.45136 & 0.43301 & 0.47133 & 0.45587 & 0.42499 & 0.4916 \\
    Deep\_Sep & 0.45482 & 0.4357 & 0.4757 & 0.46510 & 0.44477 & 0.48738 \\
    Deep\_Combine & 0.45031 & 0.45561 & 0.44513 & 0.45041 & 0.43312 & 0.46914 \\
    cls\_weighted\dag\ddag & 0.46902 & 0.41035 & 0.54725  & 0.47013 & 0.43702 & 0.50867 \\\midrule
    \textbf{SpanPro} & \textbf{0.47108} & \textbf{0.37411} & \textbf{0.63591} &  - &  - & - \\\bottomrule
    \end{tabular}%
  \label{tab:result}%
\end{table*}%

   


\begin{table}

\parbox{.45\linewidth}{
\centering
\begin{tabular}{lr}\hline
 Model & \multicolumn{1}{l}{F1} \\\hline
    LR\dag & 0.2653 \\
    emo\_BERT & 0.4781 \\
    cost\_BERT & 0.5550 \\
    BERT & 0.5795 \\
    LR\dag\ddag & 0.5475 \\
    Hybrid & 0.6369 \\
    \textbf{HybridPro} & \textbf{0.6783} \\\hline
    \end{tabular}%
    \caption{Performance of TC, \dag~ represents the inclusion of text length into the model. \ddag~  represents the inclusion of all features in Section \ref{sec:fea} into the model.}
    \label{tab:tc}
}
\hfill
\parbox{.45\linewidth}{
\centering

    \begin{tabular}{lccc}\toprule
    \multicolumn{4}{c}{SI in mini context} \\\midrule
    Model & F1 & \multicolumn{1}{c}{Precision} & \multicolumn{1}{c}{Recall} \\
    SpanBERT & {0.362} & \multicolumn{1}{c}{{0.4953}} & \multicolumn{1}{c}{{0.28523}} \\\midrule
    \multicolumn{4}{c}{TC} \\\midrule
    Model & F1 &   &  \\
    Hybrid & {0.54246} &   &  \\\bottomrule
    \end{tabular}%
    \caption{Performance of SI and TC in test dataset.}
  \label{tab:test}%

}
\end{table}

\subsection{Results of TC }
As shown in Table \ref{tab:tc}, the baseline, which only uses text length as features in Logistic Regression achieves 0.2653 of F1-measure. As for the polymorphic BERT, the emo\_BERT with extra emotion feature embedded does not obtain a better result comparing to the typical BERT. One possible explanation is that we only extract ten types of emotion feature from the lexicon, which is not sufficient for this 14 classes labelling task; in addition, many techniques use emotion appealing, so emotion is not a strong signal to distinguish one technique from another.\par
As discussed before, the training dataset is unbalanced and we refer to classes with occurrence less than 110 in training dataset as minority classes, including \textit{Whataboutism/Straw Men/Red Herring, Thought-terminating Cliches and Bandwagon/Reductio ad hitlerum}; while the rest are majority classes. In Section \ref{sec:imb}, we introduce the cost-weighted learning approach to solve the problem of imbalanced training dataset. Table \ref{tab:detail} shows that the cost-weighted learning approach outperforms on the minority classes by more than 0.20. We use the cost-weighted learning approach in our hybrid model. \par
Our experiment also shows that the Logistic Regression with the features outperforms the baseline which only integrates the text length feature, by 0.28. We compare three models: BERT, cost BERT and LR\dag\ddag with each other. For the majority classes, BERT outperforms the other two models except \textit{Repetition}, upon which LR\dag\ddag obtains improvement by 0.42. For the minority classes, cost\_BERT performs best among the three models.\par
We take advantage of each model's capacity of learning different features for different technique and integrate them as a hybrid model. Each of the submodels trains on the same whole training dataset and predicts one of the 14 classes. However, we only choose the predictions of \textit{Repetition} from LR\dag\ddag, those of other majority classes from BERT and those of minority classes from cost\_BERT. Our hybrid model outperform any its submodels by around 0.10 of F1-measure. \par
Our error analysis shows that although some instances occur only once or twice in the article, they are predicted into \textit{Repetition}. We make further efforts to correct the mislabeled classes by the rules introduced in Section \ref{sec:rule}. Our model \textbf{HybridPro} achieves 0.68 of F1-measure indicating  our rules for \textit{Repetition, Slogans, Whataboutism/Straw Men/Red Herring} are effective in TC task. The details are shown in Table \ref{tab:detail}.

\begin{table}[htbp]
  \centering
  
    \begin{tabular}{lccccc}\toprule
      Techniques & \multicolumn{1}{c}{cost\_BERT} & \multicolumn{1}{c}{BERT} & \multicolumn{1}{c}{LR} & \multicolumn{1}{c}{Hybrid} & \multicolumn{1}{c}{HybridPro} \\\midrule
    Loaded\_Language & {0.72} & {0.75} & {0.67} & {0.77} & {0.77} \\
    Name\_Calling,Labeling & {0.69} & {0.69} & {0.55} & {0.73} & {0.77} \\
    Repetition & {0.27} & {0.25} & {0.67} & {0.62} & {0.70} \\
    Doubt & {0.50} & {0.48} & {0.45} & {0.53} & {0.57} \\
    Exaggeration,Minimisation & {0.45} & {0.52} & {0.35} & {0.52} & {0.52} \\
    Appeal\_to\_fear-prejudice & {0.32} & {0.34} & {0.19} & {0.32} & {0.41} \\
    Flag-Waving & {0.70} & {0.77} & {0.69} & {0.75} & {0.79} \\
    Causal\_Oversimplification & {0.27} & {0.39} & {0.32} & {0.37} & {0.41} \\
    Appeal\_to\_Authority & {0.06} & {0.19} & {0.00} & {0.21} & {0.20} \\
    Slogans & {0.47} & {0.55} & {0.31} & {0.55} & {0.68} \\
    Black-and-White\_Fallacy & {0.13} & {0.07} & {0.12} & {0.07} & {0.14} \\
    Whataboutism,Straw\_Men,Red\_Herring & {0.18} & {0.00} & {0.07} & {0.18} & {0.24} \\
    Thought-terminating\_Cliches & {0.20} & {0.00} & {0.11} & {0.24} & {0.26} \\
    Bandwagon,Reductio\_ad\_hitlerum & {0.60} & {0.00} & {0.33} & {0.60} & {0.60} \\\bottomrule
    \end{tabular}%
    \caption{F1-measure of each technique}
  \label{tab:detail}%
\end{table}%

\subsection{Remarks}
It is worth noting that we present in Table \ref{tab:test} the performance of SpanBERT model in SI and Hybrid model in TC on the test dataset. At the test stage of the shared task, we observe that our models have the overfitting problem. We speculate that the label distribution between the development and the test datasets, hence, when the development dataset was made available again on the web site we modified the models to address the overfitting problem. The performance of our models is therefore based on the test dataset as shown in Table \ref{tab:test}. We also report here that our F1-score of 0.5340 on the development set, shown on the leaderboard for SI task, is caused by the mistake in our text pre-processing code. Specifically, we miscalculated the end index in the processing and this results in the unreasonably high recall score in the leaderboard. We fix this problem in the code and abandon that score in this paper. All the right scores are reported in this paper.
\section{Conclusion}
This paper develops a SpanBERT-based model for span identification (SI) and a hybrid model for propaganda techniques classification (TC). \par
As for SI, our paper explores different segmentation of contexts from news articles. Based on SpanBERT, we facilitate the detection by a deeper model and a sentence-level representation. The start and end boundary are conceptually independent to each other, therefore, obtaining best indexes from both the start and end classifiers achieve the best performance comparing to that of any one of them. Our model is not restricted on semantics integrity, but remaining a high ratio of span-annotated data is essential especially for a small size of training data. \par
Our experiment of TC offers several insights. First, we find that emotion features extracted from NRC lexicon is not effective to distinguish 14 classes.  Second, we find that the cost-weighted learning approach is effective in addressing the imbalance issue of the training dataset. Third, features such as text length, TF-IDF, occur times in a document, superlative form, question words, hashtags, and supplement are useful in distinguishing different propaganda techniques.\par
In the future, we will explore more on how to segment a context in the training dataset and how different context affect the results. In addition, our model lacks the ability to detect the multiple spans in one context. We will conduct a fine-grained analysis to examine whether a context contains a span and if so, how many spans are included and the exact start and end boundary of them. We also have two suggestions for future work of TC. First, the use of part of speech (POS) tags in correcting mislabeled data shows a good improvement of the performance. It would be interesting to further explore its use and representations in the model.  Second, while the hybrid model achieves improvement from sub-models, it’d be interesting to investigate a single model that differentiates 14 classes at the same time in the future.
\bibliographystyle{coling}
\bibliography{semeval2020}

\begin{thebibliography}{}

\bibitem[\protect\citename{Al-Omari \bgroup et al.\egroup
  }2019]{al2019justdeep}
Hani Al-Omari, Malak Abdullah, Ola AlTiti, and Samira Shaikh.
\newblock 2019.
\newblock Justdeep at nlp4if 2019 task 1: Propaganda detection using ensemble
  deep learning models.
\newblock In {\em Proceedings of the Second Workshop on Natural Language
  Processing for Internet Freedom: Censorship, Disinformation, and Propaganda},
  pages 113--118.

\bibitem[\protect\citename{Da~San~Martino \bgroup et al.\egroup
  }2019a]{EMNLP19DaSanMartino}
Giovanni Da~San~Martino, Seunghak Yu, Alberto Barr\'{o}n-Cede\~no, Rostislav
  Petrov, and Preslav Nakov.
\newblock 2019a.
\newblock Fine-grained analysis of propaganda in news articles.
\newblock In {\em Proceedings of the 2019 Conference on Empirical Methods in
  Natural Language Processing and the 9th International Joint Conference on
  Natural Language Processing, EMNLP-IJCNLP 2019}, EMNLP-IJCNLP 2019, Hong
  Kong, China, November.

\bibitem[\protect\citename{Da~San~Martino \bgroup et al.\egroup
  }2019b]{da2019fine}
Giovanni Da~San~Martino, Seunghak Yu, Alberto Barr{\'o}n-Cedeno, Rostislav
  Petrov, and Preslav Nakov.
\newblock 2019b.
\newblock Fine-grained analysis of propaganda in news article.
\newblock In {\em Proceedings of the 2019 Conference on Empirical Methods in
  Natural Language Processing and the 9th International Joint Conference on
  Natural Language Processing (EMNLP-IJCNLP)}, pages 5640--5650.

\bibitem[\protect\citename{Da~San~Martino \bgroup et al.\egroup
  }2020]{DaSanMartinoSemeval20task11}
Giovanni Da~San~Martino, Alberto Barr\'{o}n-Cede\~no, Henning Wachsmuth,
  Rostislav Petrov, and Preslav Nakov.
\newblock 2020.
\newblock {SemEval}-2020 task 11: Detection of propaganda techniques in news
  articles.
\newblock In {\em Proceedings of the 14th International Workshop on Semantic
  Evaluation}, SemEval 2020, Barcelona, Spain, September.

\bibitem[\protect\citename{Devlin \bgroup et al.\egroup }2018]{devlin2018bert}
Jacob Devlin, Ming-Wei Chang, Kenton Lee, and Kristina Toutanova.
\newblock 2018.
\newblock Bert: Pre-training of deep bidirectional transformers for language
  understanding.
\newblock {\em arXiv preprint arXiv:1810.04805}.

\bibitem[\protect\citename{Farkas}2018]{farkas2018disguised}
Johan Farkas.
\newblock 2018.
\newblock Disguised propaganda on social media: Addressing democratic dangers
  and solutions.
\newblock {\em Brown J. World Aff.}, 25:1.

\bibitem[\protect\citename{Gupta \bgroup et al.\egroup }2019]{gupta2019neural}
Pankaj Gupta, Khushbu Saxena, Usama Yaseen, Thomas Runkler, and Hinrich
  Sch{\"u}tze.
\newblock 2019.
\newblock Neural architectures for fine-grained propaganda detection in news.
\newblock {\em arXiv preprint arXiv:1909.06162}.

\bibitem[\protect\citename{Jones}2004]{jones2004statistical}
Karen~Sp{\"a}rck Jones.
\newblock 2004.
\newblock A statistical interpretation of term specificity and its application
  in retrieval.
\newblock {\em Journal of documentation}.

\bibitem[\protect\citename{Joshi \bgroup et al.\egroup
  }2020]{joshi2020spanbert}
Mandar Joshi, Danqi Chen, Yinhan Liu, Daniel~S Weld, Luke Zettlemoyer, and Omer
  Levy.
\newblock 2020.
\newblock Spanbert: Improving pre-training by representing and predicting
  spans.
\newblock {\em Transactions of the Association for Computational Linguistics},
  8:64--77.

\bibitem[\protect\citename{Li \bgroup et al.\egroup }2019]{li2019detection}
Jinfen Li, Zhihao Ye, and Lu~Xiao.
\newblock 2019.
\newblock Detection of propaganda using logistic regression.
\newblock In {\em Proceedings of the Second Workshop on Natural Language
  Processing for Internet Freedom: Censorship, Disinformation, and Propaganda},
  pages 119--124.

\bibitem[\protect\citename{Mohammad and Bravo-Marquez}2017]{mohammad2017wassa}
Saif~M Mohammad and Felipe Bravo-Marquez.
\newblock 2017.
\newblock Wassa-2017 shared task on emotion intensity.
\newblock {\em arXiv preprint arXiv:1708.03700}.

\bibitem[\protect\citename{Zollmann}2019]{zollmann2019bringing}
Florian Zollmann.
\newblock 2019.
\newblock Bringing propaganda back into news media studies.
\newblock {\em Critical Sociology}, 45(3):329--345.

\end{thebibliography}

\section{Appendix}
\begin{table}[htbp]
  \centering
  
    \begin{tabular}{lr}\hline
    Technique & Count \\\hline
    Loaded Language & 2,199 \\
    Name Calling,Labeling & 1,105 \\
    Repetition & 621 \\
    Doubt & 496 \\
    Exaggeration,Minimisation & 493 \\
    Appeal to fear-prejudice & 321 \\
    Flag-Waving & 250 \\
    Causal Oversimplification & 212 \\
    Appeal to Authority & 155 \\
    Slogans & 138 \\
    Black-and-White Fallacy & 112 \\
    Whataboutism,Straw Men,Red Herring & 109 \\
    Thought-terminating Cliches & 80 \\
    Bandwagon,Reductio ad hitlerum & 77 \\\hline
    \end{tabular}%
    \caption{The Count of Propaganda Technique in Training Dataset}
  \label{tab:count}%
\end{table}%
\end{document}